\documentclass{article}
\usepackage{spconf,graphicx,hyperref}
\usepackage{xcolor}
\usepackage{amsmath, amssymb, amsthm}
\usepackage{booktabs}
\usepackage{tabularx}
\usepackage{enumitem}
\usepackage{bm}

\makeatletter
\renewcommand\paragraph{\@startsection{paragraph}{4}{\z@}%
  {2ex \@plus .3ex \@minus .15ex}
  {-0.5em}
  {\normalfont\normalsize\bfseries}}
\makeatother

\title{Active Jammer Localization via Acquisition-Aware Path Planning}

\name{
Luis González-Gudiño\textsuperscript{1},
Mariona Jaramillo-Civill\textsuperscript{2},
Pau Closas\textsuperscript{2},
Tales Imbiriba\textsuperscript{1}\thanks{This work was partially supported by the National Science Foundation under Awards 1845833, 2326559 and 2530870.}}

\address{
\textsuperscript{1}Dept. of Computer Science, University of Massachusetts Boston, Boston, MA, USA \\
\textsuperscript{2}Dept. of Electrical \& Computer Engineering, Northeastern University, Boston, MA, USA
}
\begin{document}
\frenchspacing
\ninept
\maketitle
\begin{abstract}
We propose an active jammer localization framework that combines Bayesian optimization with acquisition-aware path planning. Unlike passive crowdsourced methods, our approach adaptively guides a mobile agent to collect high-utility Received Signal Strength measurements while accounting for urban obstacles and mobility constraints. For this, we modified the A* algorithm, A-UCB*, by incorporating acquisition values into trajectory costs, leading to high-acquisition planned paths. Simulations on realistic urban scenarios show that the proposed method achieves accurate localization with fewer measurements compared to uninformed baselines, demonstrating consistent performance under different environments.
\end{abstract}
\section{Introduction}
\label{sec:intro}

Global Navigation Satellite Systems (GNSS) such as GPS, Galileo, GLONASS and BeiDou provide critical position, navigation, and timing (PNT) services for a wide array of applications, from intelligent transportation and precision agriculture to timing-dependent infrastructures like banking systems and cellular networks \cite{PNTbook}. However, the strong dependence on GNSS makes these systems vulnerable to both unintentional and intentional interference \cite{VulnerabilitiesArticle, VulnerabilitiesArticle2}. Unintentional interference may originate from out-of-band sources such as terrestrial digital video broadcasting (DVB-T) or amateur radios, as well as from in-band sources like distance measuring equipment (DME) or civilian radars. Additionally, GNSS signals are susceptible to jamming by personal privacy devices (PPDs). These devices, which are inexpensive and readily available online despite being illegal, emit high-power signals in the L-band (the frequency band used by GNSS) and can disrupt reception over distances ranging from tens of meters to several kilometers. These interferences can overwhelm receiver front-ends and result in service denial, with incidents reported in ports and air traffic control zones \cite{exelisReport}. 

Detecting and localizing such interferers is essential for resilient PNT operations. A cost-effective and scalable solution is crowdsourced data \cite{crowdsourcing1, APBM, crowdsourcing2}, especially in densely populated or high-traffic areas with many GNSS users. Crowdsourcing leverages existing GNSS-enabled devices (e.g., smartphones) to collect environmental data in a distributed way. A key measurement here is Received Signal Strength (RSS), which quantifies signal power at a given location and frequency. Under jamming, RSS can deviate markedly from expected GNSS levels, indicating abnormal activity. By analyzing spatially distributed RSS readings, it becomes possible to infer the presence and even approximate location of interference sources without requiring a dense deployment of dedicated monitoring stations.

This principle has given rise to a variety of localization techniques. Many approaches in crowdsourced jammer localization typically rely on fitting power measurements, such as Carrier-to-Noise-density ratio $C/N_0$ or Automatic Gain Control (AGC) values, to a simple physical model \cite{crowdsourcing2, localization1, localization2, localization3, GNSSdata}. These methods often assume a known path-loss propagation model that, while effective in open-sky scenarios, struggles in complex urban environments where multipath, shadowing and occlusions introduce significant deviations from the ideal path-loss function. To address these limitations, more recent work has explored data-driven approaches. Instead of assuming a fixed physical model, these methods use tools like neural networks to learn the complex, non-linear relationship between location and RSS directly from the data \cite{APBM, jaramillo}.

A common thread in the aforementioned research is its passive nature. These methods rely on data collected by users pursuing their own objectives, meaning samples are gathered incidentally. This leads to inefficient localization, as measurements may be sparse, clustered in redundant areas, or fail to cover regions of highest uncertainty. As a result, many samples may be needed for a confident jammer estimate. Several recent works have explored alternative approaches to overcome these limitations. For instance, \cite{active1} propose a UAV-based system that scans the environment by hovering at preplanned waypoints and rotating a directional antenna to measure signal strength. However, the UAV follows a static plan and does not adapt its path to the data, limiting efficiency. In contrast, \cite{active2} deploy a UAV tailored for jamming scenarios that performs greedy bearing-based triangulation. This adds partial adaptivity by iteratively steering toward estimated jammer directions, but remains limited to short-horizon heuristics without global reasoning about promising regions.

These limitations reveal a gap in the literature: the absence of adaptive strategies that guide data collection in a sample-efficient and environment-aware manner. To address this, we propose an active localization framework that combines Bayesian optimization with acquisition-aware path planning, enabling an autonomous agent to navigate complex environments while wisely selecting high-value measurements. To isolate the performance of our proposed framework, we focus on the localization of a single static jammer. However, with an appropriately designed multimodal surrogate model, the framework could be extended to handle multiple jammers. In summary, our main contributions are: 

\begin{itemize}
    \item A novel Bayesian optimization framework for active jammer localization.
    \item An acquisition-aware path planning strategy that balances movement cost and acquisition gain.
    \item A sample-efficient strategy that accurately localizes the jammer with minimal measurements.
\end{itemize}


\section{Problem Formulation}
\label{sec:problem}
 
In this work, we address the problem of localizing a single stationary jamming source in urban environments, using crowdsourced RSS measurements from static agents together with adaptive active sensing by an autonomous mobile agent. Our key assumption is that the jammer’s true position $\mathbf{x}_J$ lies in a bounded two-dimensional area $\mathcal{X}\subset\mathbb{R}^2$  and induces an (unknown) interference-power field $f_{\rm true}(\mathbf{x};\mathbf{x}_J)$, whose global maximum corresponds to the location where the RSS from the jammer is strongest. Thus, our goal is to estimate it by finding the global maximizer of this field:

\begin{equation}
\hat{\mathbf{x}}_J = \arg\max_{\mathbf{x} \in \mathcal{X}} f_{\rm true}(\mathbf{x}; \mathbf{x}_J)
\label{eq:xjamopt}
\end{equation}

Obtaining a closed-form analytical model for $f_{\rm true}$ in urban scenarios is extremely challenging due to multipath, non-line-of-sight conditions, and shadowing effects that make the field highly non-convex and dependent on city-specific characteristics like building layout and materials. As a consequence, $f_{\rm true}$ is treated as a black-box function. In this context, an autonomous agent is deployed to sequentially visit probing locations can collect data with the ultimate goal of solving~\eqref{eq:xjamopt}. At each location $\mathbf{x} \in \mathcal{X}$, the agent records a noisy RSS measurement:

\begin{equation}
y_n = f_{\rm true}(\mathbf{x}\,; \mathbf{x}_J) + \xi_n, \quad n=1, \dots, N
\label{eq:measurement}
\end{equation}

\noindent with $\xi_n$ representing additive measurement noise. Importantly, the agent operates under navigation constraints: some regions of $\mathcal{X}$ are inaccessible due to static obstacles such as buildings, walls, or restricted zones. We assume that these obstacles (and hence the feasible subset of $\mathcal{X}$) are known a priori. Therefore, the challenge is not only to search for the global maximizer, but to decide where to measure next so that the maximizer can be identified with as few samples as possible. This motivates an approach that smartly balances exploration and exploitation.

\section{Framework for Jammer Localization}
\label{sec:model}
To tackle this problem effectively, we require a strategy capable of navigating complex, noisy, and partially observable environments in a data-efficient manner. Bayesian optimization (BO) provides a natural fit for this setting: it is specifically designed for optimizing expensive, black-box functions with limited samples, while explicitly modeling uncertainty. In our context, BO enables the agent to reason about both the expected interference power and the confidence of that estimate across the environment, allowing it to actively trade off, through an acquisition function, between exploring uncertain regions and exploiting promising ones. For a detailed overview of the BO framework, we refer the reader to \cite{brochu2010}.

\begin{figure}
    \centering
    \includegraphics[width=\linewidth]{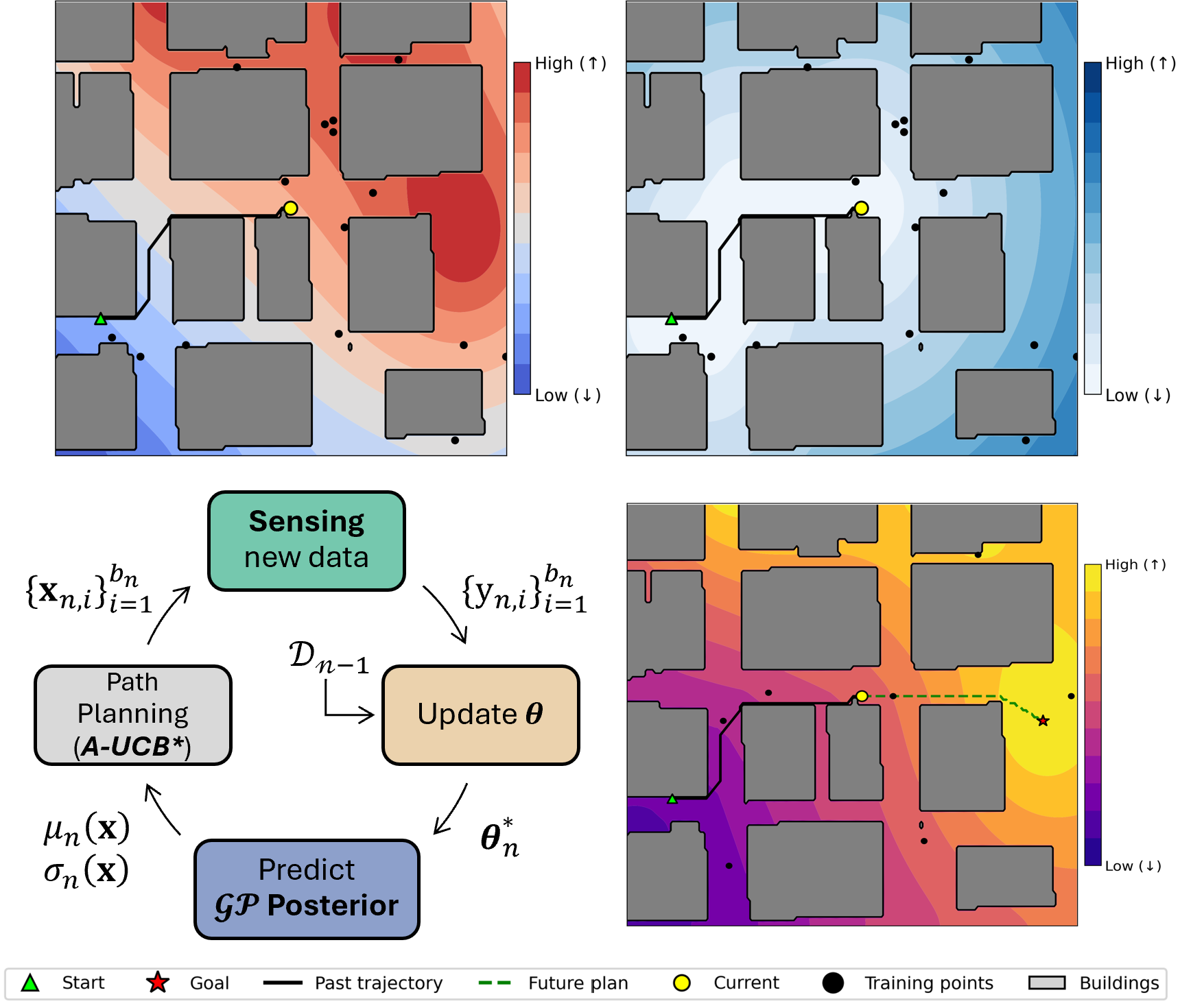}
    \caption{Overview of the proposed active localization framework. \textbf{Top:} GP posterior mean $\mu_n(\mathbf{x})$ and uncertainty $\sigma_n(\mathbf{x})$ over $\mathcal{F}$. \textbf{Bottom-right:} UCB acquisition $\alpha_{\mathrm{UCB}}(\mathbf{x})$ used to select the next sensing target $\mathbf{x}^{c}_{n}$ ($\star$). The black curve is the past trajectory and the green dashed line is the path planning toward $\mathbf{x}^{c}_{n}$. Gray polygons are buildings (non-traversable) and black dots are initial crowdsourced samples $\mathcal{D}_0$.}
    \label{fig:framework}
\end{figure}

Inspired by BO, we adopt an iterative framework (see Fig.\ref{fig:framework}) where at each iteration, the agent performs four main steps: (i) acquires a new measurement at a selected location, (ii) updates a probabilistic model of the interference field, (iii) computes an acquisition function that determines the next sensing location, and (iv) plans an acquisition-guided path to this location. This loop continues until convergence criteria are met, such as reaching a confidence in the estimated jammer position or a time constraint.

In our approach, we discretize the continuous domain $\mathcal{X} \subset \mathbb{R}^2$ into a finite set of uniformly spaced grid points denoted as $\mathcal{G} = \{\mathbf{x}^{(1)}, \dots, \mathbf{x}^{(M)}\} \subset \mathcal{X}$, where each $\mathbf{x}^{(i)}$ corresponds to a grid cell. For notational simplicity, we will write $\mathbf{x}^{(i)}$ simply as $\mathbf{x}$. This grid-based representation serves two purposes: first, it simplifies the fitting of the surrogate model and acquisition function evaluation, and second, it enables efficient path planning via graph-based algorithms. Some cells in $\mathcal{G}$ are obstructed by static obstacles. Let each individual obstacle be defined as a subset $\mathcal{O}_k \subset \mathcal{G}$ for $k = 1, \dots, K$, and define the complete set of obstacle cells as $\mathcal{O} = \bigcup_{k=1}^{K} \mathcal{O}_k \subset \mathcal{G}$. Therefore, the feasible set of grid locations where the agent can safely navigate and take measurements is then given by $\mathcal{F} = \mathcal{G} \setminus \mathcal{O}$. All agent trajectories and sensing decisions are constrained to lie within this set. We now describe each component of the framework in more detail.

\paragraph*{Sensing.} At each iteration $n = 1, \dots, N$ of the BO loop, the agent selects a batch of $b_n$ feasible locations $\{\mathbf{x}_{n,1}, \dots \mathbf{x}_{n,b_n}\} \subset \mathcal{F}$ (see Sec.\ref{sec:aucbstar}) and records a corresponding set of noisy RSS measurements according to~\eqref{eq:measurement}. In this article, noise is assumed to be Gaussian $\xi_n \sim \mathcal{N}(0, \sigma^2)$. These new measurements are appended to the existing dataset, which evolves over time as $\mathcal{D}_n = \mathcal{D}_{n-1} \cup \{ (\mathbf{x}_{n,j}\,, y_{n,j})\}_{j=1}^{b_n}$. Here, $\mathcal{D}_0$ contains $b_0$ initial crowdsourced measurements available at time step zero, each also obtained according to~\eqref{eq:measurement} by static agents.


\paragraph*{Predictive Model.} 

We adopt a \emph{Gaussian Process} (GP) \cite{gp} regressor as surrogate model of the interference-power field. GPs provide a nonparametric Bayesian prior over functions that yields, after conditioning on data, closed-form posterior predictive means and variances. In short, we assume the prior $f_\text{surr}(\tilde{\mathbf{x}}) \sim \mathcal{GP}\big(0,\,k_\theta(\tilde{\mathbf{x}},\tilde{\mathbf{x}}')\big),$ where at each location $\mathbf{x} = (p_x, p_y)$ we build the feature vector $\tilde{\mathbf{x}} = [p_x,\, p_y,\, z_\mathbf{x}],$
with $p_x,p_y$ the normalized 2D coordinates and $z_\mathbf{x}$ the normalized building height at location $\mathbf{x}$. This lightweight augmentation helps the GP capture systematic power variations induced by urban morphology without sampling inside obstacles.

Given the dataset $\mathcal{D}_n$, the surrogate model follows a posterior distribution $f_{\rm surr,n}|\mathcal{D}_n,\tilde{\mathbf{x}} 
\sim \mathcal{N} (\mu_n(\tilde{\mathbf{x}}), \sigma_n^2(\tilde{\mathbf{x}}))$ at any query location $\tilde{\mathbf{x}}$ with mean $\mu(\tilde{\mathbf{x}})$ and uncertainty $\sigma(\tilde{\mathbf{x}})$:

{
\abovedisplayskip=2pt
\abovedisplayshortskip=2pt
\begin{align}
\mu_n(\tilde{\mathbf{x}}) 
    &= \mathbf{k}_{\theta,n}(\tilde{\mathbf{x}})^\top \big(\mathbf{K}_n+\sigma_\eta^2 I\big)^{-1}\mathbf{y}_n\\
\sigma_n^2(\tilde{\mathbf{x}}) 
    &= k_\theta(\tilde{\mathbf{x}},\tilde{\mathbf{x}}) - 
       \mathbf{k}_{\theta,n}(\tilde{\mathbf{x}})^\top \big(\mathbf{K}_n+\sigma_\eta^2 I\big)^{-1}\mathbf{k}_{\theta,n}(\tilde{\mathbf{x}})
\end{align}
}

\noindent where $\mathbf{y}_n=[y_1,\dots,y_{|\mathcal D_n|}]^\top$ denotes the observation vector and $\mathbf{K}_n \in \mathbb{R}^{|\mathcal D_n|\times |\mathcal D_n|}$ defines the Gram matrix  with entries $[\mathbf{K}_n]_{pq}=k_\theta(\tilde{\mathbf{x}}_p,\tilde{\mathbf{x}}_q')$. For a new query $\tilde{\mathbf{x}}$, the associated kernel vector is $\mathbf{k}_{\theta,n}(\tilde{\mathbf{x}})=[k_\theta(\tilde{\mathbf{x}},\tilde{\mathbf{x}}_1),\dots,k_\theta(\tilde{\mathbf{x}},\tilde{\mathbf{x}}_{|\mathcal D_n|})]^\top$. These two quantities are jointly exploited by the acquisition function (see Sec.~\ref{sec:acquistion}). To flexibly capture both short-range fluctuations (e.g., local multipath) and broader trends of the field, we use an additive, multi-scale kernel with a learnable noise term:
\begin{align}
k_\theta(\tilde{\mathbf{x}},\tilde{\mathbf{x}}') = k^\ell_\theta(\tilde{\mathbf{x}},\tilde{\mathbf{x}}') + k^s_\theta(\tilde{\mathbf{x}},\tilde{\mathbf{x}}') + \sigma_\eta^2 \delta_{\tilde{\mathbf x}, \tilde{\mathbf x}'}
\label{eq:kernel}
\end{align}
with $k^\iota_\theta(\tilde{\mathbf{x}},\tilde{\mathbf{x}}') = \sigma_\iota^2 \exp(-\tfrac{1}{2}\!\sum_{d} {\tilde{\mathbf{x}}-\tilde{\mathbf{x}}')^2}/{\ell_{\iota,d}^2})$, and $\iota\in\{\ell, s\}$ corresponding to the long- and short-length scales, explaining the smooth and high-frequency components of the field. The white-noise term $\sigma_\eta^2\delta_{\tilde{\mathbf x}, \tilde{\mathbf x}'}$ accounts for measurement noise and residual model error, with $\delta_{\tilde{\mathbf x}, \tilde{\mathbf x}'}=1$ if $\tilde{\mathbf x} = \tilde{\mathbf x}'$ and 0 otherwise. We found this two-scale structure a good compromise between expressiveness and robustness for urban fields. Our formulation is agnostic to this choice and can swap kernels without altering the rest of the pipeline.

Hyperparameters $\bm{\theta}=(\sigma_s^2,\sigma_\ell^2,\ell_{s,d},\ell_{\ell,d},\sigma_\eta^2)$ are learned by maximizing the GP log marginal likelihood with multiple random restarts. Moreover, targets are normalized (zero-mean, unit-variance) and de-normalized at prediction time. Given $\mathcal{D}_n$, the posterior predictive is computed in closed form and supplied to the acquisition (see \ref{sec:acquistion}). Finally, the active sensing framework does \textit{not} rely on GP-specific structure: any probabilistic surrogate able to produce calibrated uncertainty (e.g., Bayesian neural networks) could replace the GP. We choose GPs here for their data efficiency in the low-to-moderate sample regime and their well-calibrated uncertainties, which are pivotal for acquisition-aware planning.

\paragraph*{Acquisition Function.}To select the next candidate location, we adopt the Upper Confidence Bound (UCB) \cite{ucb} acquisition strategy. UCB selects points that maximize a weighted combination of the surrogate model's predictive mean and standard deviation:
\label{sec:acquistion}
\begin{equation}
\alpha_{\text{UCB}}(\mathbf{x}) = \mu(\mathbf{x}) + \kappa \cdot \sigma(\mathbf{x})
\end{equation}

\noindent where $\mu(\mathbf{x})$ and $\sigma(\mathbf{x})$ are the predictive mean and standard deviation at location $\mathbf{x}$, and $\kappa > 0$ is a tunable parameter that controls the exploration–exploitation trade-off. Higher values of $\kappa$ encourage more exploration by favoring regions with high uncertainty, while lower values bias the search toward locations with high predicted signal strength. At each iteration, the next candidate point is chosen as $\mathbf{x}_{n}^{c} = \arg\max_{\mathbf{x} \in \mathcal{F}} \, \alpha_{\text{UCB}}(\mathbf{x})$.

\paragraph*{Path Planning Strategy.}While Bayesian optimization identifies the next most valuable location $\mathbf{x}_{n}^{c}$ via an acquisition function, navigating directly to this target is often infeasible or suboptimal in real environments. Urban landscapes impose hard constraints due to buildings, and moving in straight lines can lead to redundant paths that fail to extract new information. What is needed is a path planning strategy that remains goal-oriented but also exploits the acquisition landscape along the way. In other words, at each BO iteration we set $\mathbf{x}_{n}^{c}$ as the intended destination, and instead of moving directly there, we compute a bounded-length path from the agent's current position that also prioritizes acquisition gain.\label{sec:aucbstar}

To this end, we modify the A* algorithm \cite{astar1, astar2}, a graph-based search method for computing minimum-cost paths. For each node $\mathbf{x}$, the algorithm evaluates  
\begin{equation}
    f(\mathbf{x}) = g(\mathbf{x}) + h(\mathbf{x}),
\end{equation}  
where $g(\mathbf{x})$ is the accumulated path cost from the start node (sum of edge costs $c_{\mathbf{x}, \mathbf{x}'}$) and $h(\mathbf{x})$ is an admissible heuristic that estimates the remaining cost to the goal. By always expanding the node with minimum $f(\mathbf{x})$, A* guarantees an optimal collision-free path in grid environments with static obstacles. While classical A* minimizes only travel cost, yielding efficient but uninformed paths, we modify the arc cost definition to incorporate values from the acquisition function. Specifically, for each edge connecting nodes $\mathbf{x}$ and $\mathbf{x}'$, we define  
\begin{equation}
    c_{\mathbf{x},\mathbf{x}'} = \big(\lambda_{\text{len}} - \lambda_{\text{info}} \cdot \bar{\alpha}\big) \, \|\mathbf{x} - \mathbf{x}'\|,
\end{equation}  
where $\bar{\alpha} = \tfrac{1}{2} \big(\alpha_{\text{UCB}}(\mathbf{x}) + \alpha_{\text{UCB}}(\mathbf{x}')\big)$ is the mean acquisition value across the edge. Here, $\lambda_{\text{len}}$ weighs travel cost, while $\lambda_{\text{info}}$ biases the planner toward promising regions. This modification transforms A* into an acquisition-aware path planner A-UCB* that not only reaches the selected target $\mathbf{x}^{c}_{n}$ but also collects valuable samples along the way. To respect mobility constraints, each path is limited to a maximum length budget $\delta$ (implicitly accounting for a maximum allowable velocity) and uniformly subsampled into $b_n$ waypoints per iteration.  

\vspace{-0.3cm}
\section{Experiments}

\subsection{Data Generation}
\label{sec:data}

We synthesize RSS fields following the crowdsourced-jamming setups in prior work \cite{APBM, jaramillo}, using MATLAB’s deterministic 3D ray-tracing engine \cite{raytracing1, raytracing2}. Unlike prior work, we adopt a denser sampling strategy, placing receivers on a uniform grid with $2$\,m spacing between adjacent sampling points across the walkable workspace. Building maps are also preprocessed: interior courtyards without street access are removed so that each building becomes a compact obstacle polygon. We considered two representative urban layouts:
\begin{itemize}
    \item \textbf{Chicago Downtown (dense urban core):} a downtown scenario characterized by narrow streets and tall buildings, creating strong multipath and shadowing effects.  
    \item \textbf{Boston Common (urban park with open-sky):} a mixed environment where open park areas provide long line-of-sight corridors, while surrounding façades still generate significant reflections and occlusions.
\end{itemize}

\subsection{Quantitative Results}
\label{sec:quant_results}

We evaluate the performance of our jammer localization framework under various sampling andplanning strategies. The goal of this analysis is dual: first, to assess how accurately each method can localize the jammer using a limited number of iterations, and second, to investigate the role of acquisition-aware path planning in improving sample efficiency.

Besides our method with a finite path budget, A-UCB* $(\delta=50)$, we consider: (i) A-UCB* with unlimited path length $(\delta=\infty)$ as an upper bound; (ii) Random Motion (RM), which moves uniformly at random along the four cardinal directions with $\delta=50$ steps per iteration—this respects motion feasibility but ignores acquisition information, serving as a conservative lower bound; and (iii) Random i.i.d. Sampling (RIS), which draws queries uniformly from $\mathcal{F}$ without motion continuity or acquisition awareness.


All methods are evaluated over $100$ independent trials. At each trial, the agent starts from a randomly selected initial position in the grid; this position is fixed across all methods within the same trial to ensure a fair comparison. In every BO iteration, all methods collect $b_n = 2$ RSS measurements and all experiments use $b_0 = 35$ initial crowdsourced  samples uniformly drawn from $\mathcal{F}$. The UCB exploration parameter is set to $\kappa = 2$, and the measurement noise variance to $\sigma^2 = 2.5$. For A-UCB*, the maximum path length budget is set to $\delta = 50$. We fixed these values for all methods and trials.

\begin{figure}[h!]
    \centering
    \includegraphics[width=0.85\columnwidth]{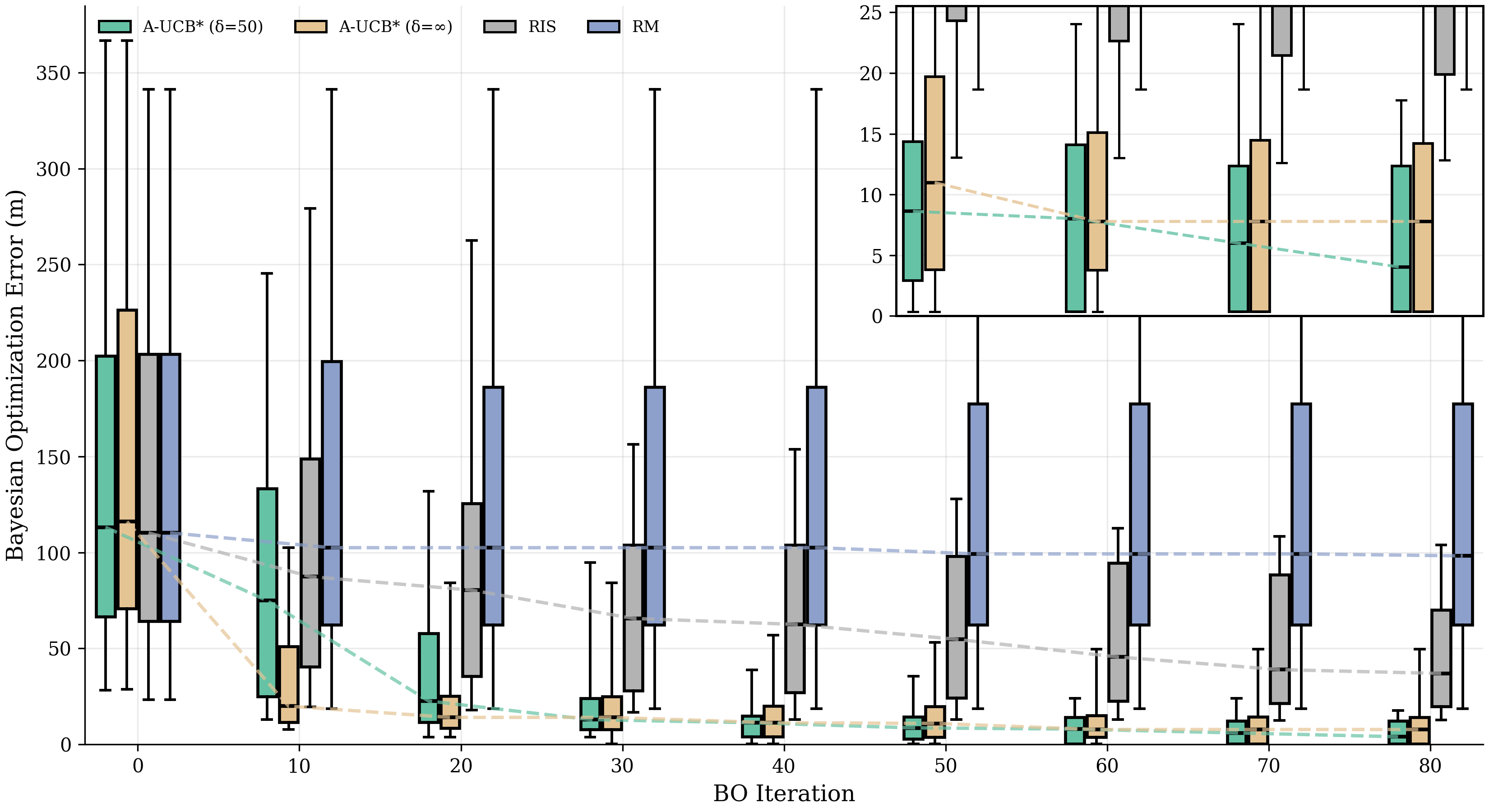}\\
    \includegraphics[width=0.85\columnwidth]{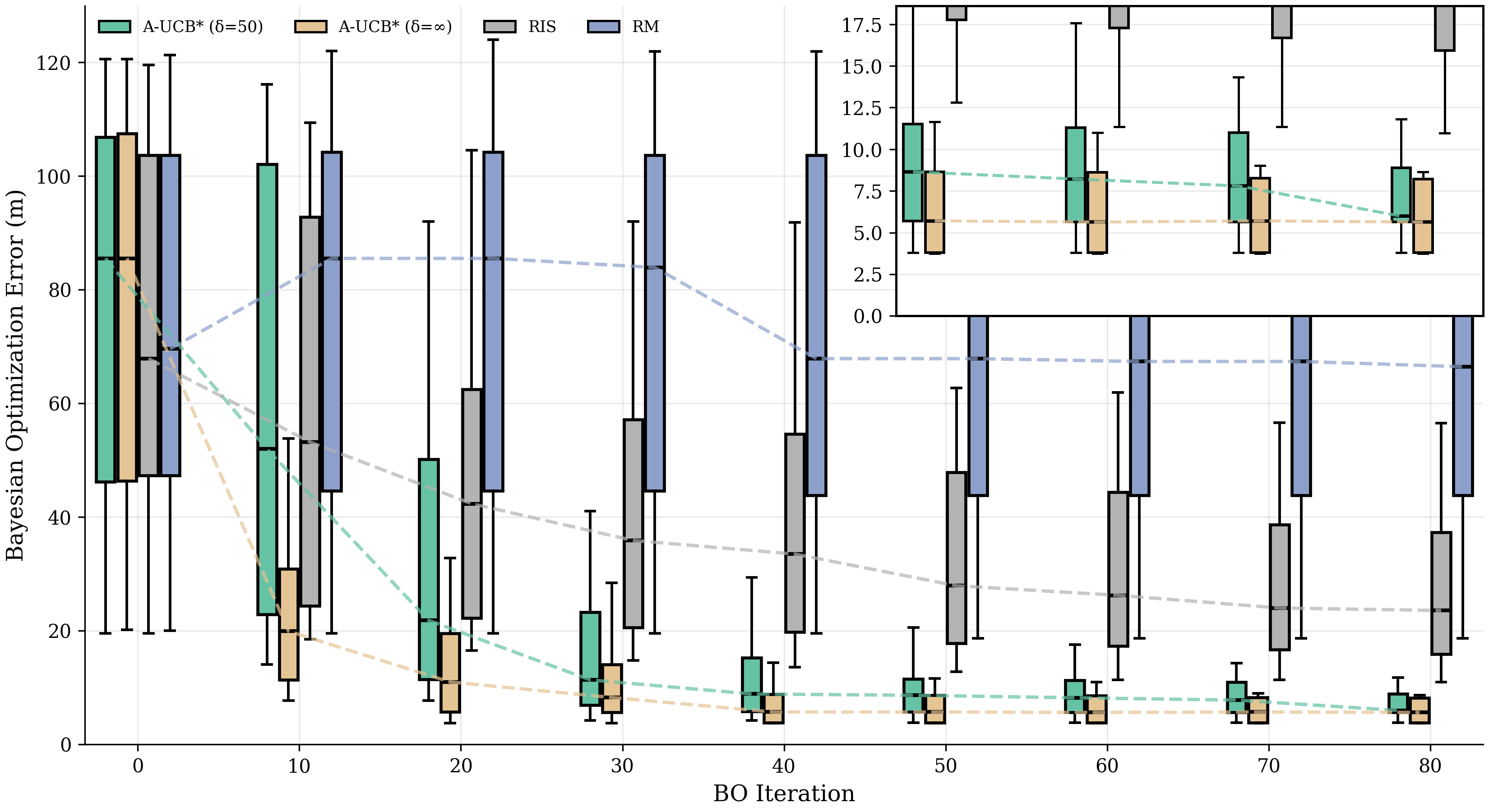}
    \caption{Evolution of the BOE in the \textbf{Chicago Downtown (top panel)} and \textbf{Boston Common (bottom panel)} scenarios across 80 Bayesian optimization iterations. Boxplots at each iteration show the distribution of localization errors across 100 independent trials. The inset zooms into the last 30 iterations highlight last convergence behavior.}
    \label{fig:Chicago_boxplots_bestpoint}
\end{figure}

To evaluate localization accuracy, we report two complementary error metrics. The Surrogate Model Error (SME) measures the distance between the true jammer location and the maximizer of the GP posterior mean field, reflecting the accuracy of the surrogate prediction. The Bayesian Optimization Error (BOE) instead measures the distance to the grid point with the highest RSS value actually sampled, which corresponds to the output of the proposed algorithm. The results are summarized in Table \ref{tab:quant_results}, where we report the median and interquartile range (25\%-75\%) of each error metric across the $100$ trials. Furthermore, we illustrate the evolution of the BOE across the 80 iterations in Fig.\ref{fig:Chicago_boxplots_bestpoint}. Specifically, these plots depict the distribution of BOE values at each iteration for both scenarios.

\begingroup
\setlength{\intextsep}{5pt}          
\setlength{\textfloatsep}{6pt}       
\setlength{\abovecaptionskip}{0pt}   
\setlength{\belowcaptionskip}{0pt}   
\begin{table}[h!]
\centering
\caption{Localization error (median [25\%–75\%]) across $100$ independent trials for both Chicago Downtown and Boston Common datasets. Errors are reported in meters for SME and BOE. Lower ($\downarrow$) values indicate better localization performance.}
\label{tab:quant_results}
\resizebox{\columnwidth}{!}{
\renewcommand{\arraystretch}{1.4} 
\begin{tabular}{lcccc}
\toprule
& \multicolumn{2}{c}{\textbf{Chicago Downtown}} & \multicolumn{2}{c}{\textbf{Boston Common}} \\
\cmidrule(lr){2-3} \cmidrule(lr){4-5}
\textbf{Method} & \textbf{SME} ($\downarrow$) & \textbf{BOE} ($\downarrow$) & 
\textbf{SME} ($\downarrow$) & \textbf{BOE} ($\downarrow$) \\
\midrule
RIS & 57.4 [27.6–120.1] & 65.6 [28.1–103.8] & 50.7 [25.4–93.6] & 35.9 [20.6–57.2] \\
RM  & 100.3 [36.9–226.5] & 102.5 [62.3–186.1] & 82.4 [46.0–132.4] & 83.9 [44.6–103.6] \\
\midrule
\textbf{A-UCB*} $(\delta = \infty)$ & 14.1 [6.5–24.5] & 14.1 [7.7–25.0] & 8.3 [5.3–12.4] & 8.3 [5.6–14.1] \\
\textbf{A-UCB*} $(\delta = 50)$     & 13.4 [7.8–28.0] & 12.9 [7.8–24.0] & 12.8 [8.3–32.2] & 11.4 [6.9–23.2] \\
\bottomrule
\end{tabular}
}
\end{table}
\endgroup

The proposed acquisition-aware planning strategy consistently achieves faster convergence and lower localization errors compared to the uninformed baselines. Random motion is particularly ineffective, as its tendency to revisit redundant regions leads to persistent errors even after many iterations. Uniform random sampling performs slightly better due to broader spatial coverage, but it still lacks the adaptivity needed to consistently reduce error. In contrast, A-UCB* rapidly drives the optimization process toward the jammer location, with its error distribution narrowing significantly after about 30 iterations—corresponding to fewer than 100 total measurements when accounting for the initial 35 crowdsourced samples and two agent-collected samples per iteration. This outcome underscores the effectiveness of acquisition-aware path planning in prioritizing high-utility measurements: rather than relying on exhaustive exploration, the method focuses on high-value measurements, producing more reliable field estimates minimizing wasted effort. Moreover, even when constrained by a finite path-length budget, the method retains most of its efficiency. Finally, the comparison between the two datasets emphasizes the role of the environment: Boston Common, with its open-sky areas, yields lower error levels overall, while the dense and obstructed Chicago Downtown setting poses greater challenges. Despite this, A-UCB* maintains robust performance in both scenarios, highlighting its adaptability in complex urban environments.

\vspace{-0.1cm}
\subsection{Sensitivity of \texorpdfstring{$\kappa$}{kappa}}
\label{sec:kappa_sensitivity}

\begin{figure}[h!]
    \centering
    \includegraphics[width=0.85\columnwidth]{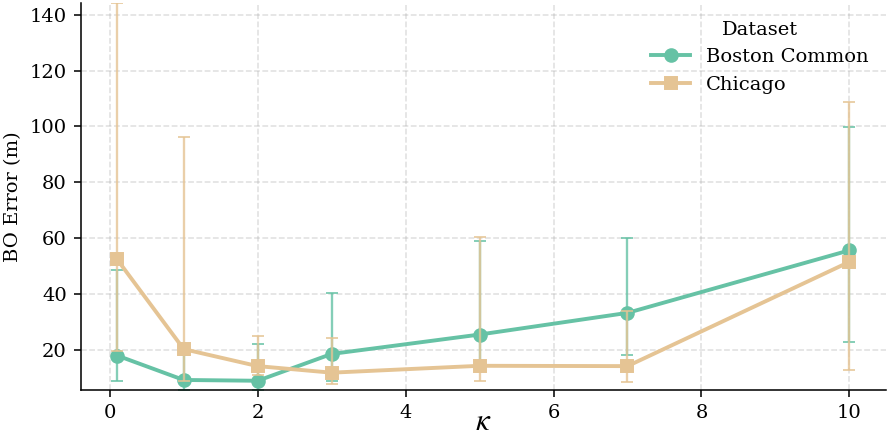}\vspace{-8pt}
    \caption{Sensitivity analysis of the exploration–exploitation parameter $\kappa$ in the UCB acquisition function. The plot shows the BOE after 30 iterations for both datasets. Each point corresponds to the median BOE across 100 trials, with error bars denoting the IQR.}
    \label{fig:kappa}
\end{figure}

Varying $\kappa$ and measuring BOE at iteration 30 shows a shallow U-shaped trend in both datasets, as illustrated in Fig.~\ref{fig:kappa}. Very small $\kappa$ (e.g., 0.1) is too exploitative, yielding large median errors and wide IQRs, where the agent’s initialization strongly influences performance. Very large $\kappa$ ($\geq 5$) over-explores, increasing error and variability at the fixed number of iterations. The lowest, most stable errors occur for $\kappa \in [1,3]$. We therefore find $\kappa=2$ a robust choice within the low-error plateau.

\vspace{-5mm}

\section{Conclusion}
We introduced an acquisition-aware Bayesian optimization framework for jammer localization that integrates probabilistic modeling with adaptive path planning. Results in realistic urban environments confirm that our method localizes interference sources with high accuracy using limited measurements, outperforming random baselines in both convergence speed and robustness. Even under mobility constraints, the framework retains efficiency, highlighting its suitability for real-world deployments. Future work will extend this approach to multi-jammer scenarios and explore alternative probabilistic surrogates beyond GP's.



\vfill\pagebreak
\clearpage

\bibliographystyle{IEEEbib}
\bibliography{bibliography}

\end{document}